\documentclass[runningheads]{llncs}
\usepackage[T1]{fontenc}
\usepackage[nolist,nohyperlinks]{acronym}
\usepackage[export]{adjustbox}
\usepackage{multirow}
\usepackage{makecell}
\usepackage{subfigure}
\usepackage{float}
\usepackage{url}

\usepackage{amsmath}
\usepackage{amssymb}
\usepackage{mathtools}

\usepackage{bm}
\usepackage{booktabs}

\usepackage{siunitx,etoolbox}

\usepackage{graphicx}
\begin{document}
\sloppy
\begin{acronym}
\acro{CXR}{chest x-ray}	
\acro{CNN}{convolutional neural network}	
\acro{LSFM}{last spatial feature map}		
\acro{ET}{eye-tracking}	
\acro{AUC}{area under the receiver operating characteristic curve}	
\acro{GAP}{global average pooling}	
\acro{CAM}{class activation map}
\acro{NCC}{normalized cross-correlation}
\end{acronym}

\title{Comparing radiologists' gaze and saliency maps generated by interpretability methods for chest x-rays}
\titlerunning{Comparing radiologists' gaze and saliency maps for chest x-rays}

%
%

\author{Ricardo Bigolin Lanfredi\inst{1}\orcidID{0000-0001-8740-5796} \and
Ambuj Arora\inst{2}\orcidID{0000-0001-7740-7333} \and
Trafton Drew\inst{3} \and
Joyce D. Schroeder\orcidID{0000-0002-7451-4886} \and
Tolga Tasdizen\inst{1}\orcidID{0000-0001-6574-0366}}


\authorrunning{R. Bigolin Lanfredi et al.}

%
\institute{Scientific Computing and Imaging Institute, University of Utah, Salt Lake City, UT, USA \\ \email{ricbl@sci.utah.edu} \and
School of Computing, University of Utah, Salt Lake City, UT, USA \and
Department of Psychology, University of Utah, Salt Lake City, UT, USA \and Department of Radiology and Imaging Sciences, University of Utah, Salt Lake City, UT, USA}


%
\maketitle              
\begin{abstract}
The interpretability of medical image analysis models is considered a key research field. We use a dataset of eye-tracking data from five radiologists to compare the outputs of interpretability methods and the heatmaps representing where radiologists looked. We conduct a class-independent analysis of the saliency maps generated by two methods selected from the literature: Grad-CAM and attention maps from an attention-gated model. For the comparison, we use shuffled metrics, which avoid biases from fixation locations. We achieve scores comparable to an interobserver baseline in one shuffled metric, highlighting the potential of saliency maps from Grad-CAM to mimic a radiologist's attention over an image. We also divide the dataset into subsets to evaluate in which cases similarities are higher.

\keywords{Interpretability \and XAI \and Chest X-rays \and Radiology \and Eye Tracking \and Gaze \and Saliency Maps}
\end{abstract}

\section{Introduction}

The interpretability of deep learning models is an essential property for their adoption in the medical field~\cite{key}. Several interpretability techniques produce saliency maps highlighting the spatial areas of most importance for each of the network's outputs, such as Grad-CAM~\cite{gradcam} and spatial attention maps~\cite{ag}. We propose to quantify\footnote{Code is available at
\url{https://github.com/ricbl/etsaliencymaps} } 
the similarity between human attention and the saliency maps produced by these methods.

\begin{figure}[ht]

\centerline{\centering \adjustbox{minipage=0.85\linewidth,valign=t}{\begin{subfigure}[]{\label{main:a}\includegraphics[width=0.205\linewidth,valign=t]{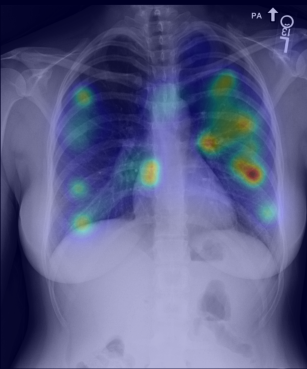}}
  \end{subfigure}
  \begin{subfigure}[]{\label{main:b}\includegraphics[width=.205\linewidth,valign=t]{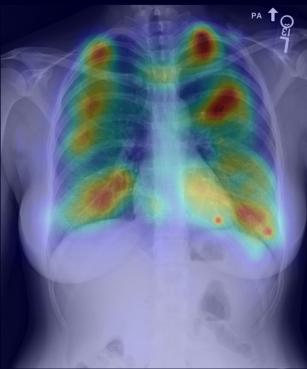}}
  \end{subfigure}
    \begin{subfigure}[]{\label{main:c}\includegraphics[width=.205\linewidth,valign=t]{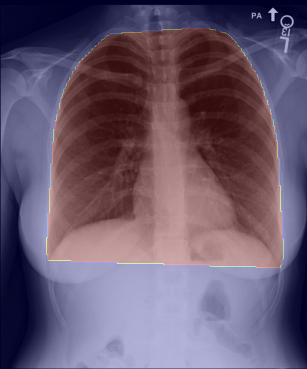}}
  \end{subfigure}
    \begin{subfigure}[]{\label{main:d}\includegraphics[width=.205\linewidth,valign=t]{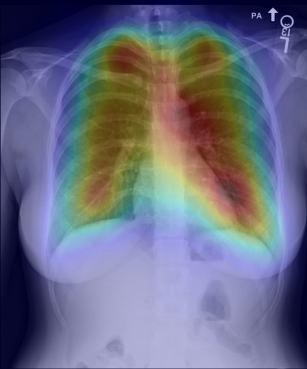}}
  \end{subfigure}
  \vskip -0.5em
    \begin{subfigure}[]{\label{main:e}\includegraphics[width=.205\linewidth,valign=t]{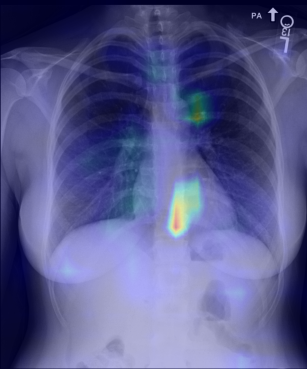}}
  \end{subfigure}
    \begin{subfigure}[]{\label{main:f}\includegraphics[width=.205\linewidth,valign=t]{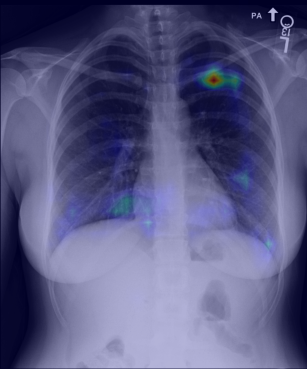}}
  \end{subfigure}
    \begin{subfigure}[]{\label{main:g}\includegraphics[width=.205\linewidth,valign=t]{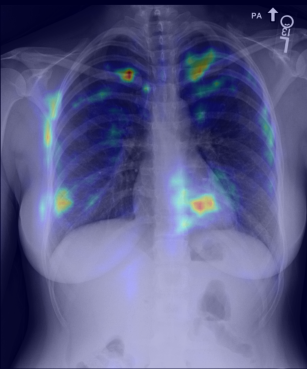}}
  \end{subfigure}
    \begin{subfigure}[]{\label{main:h}\includegraphics[width=.205\linewidth,valign=t]{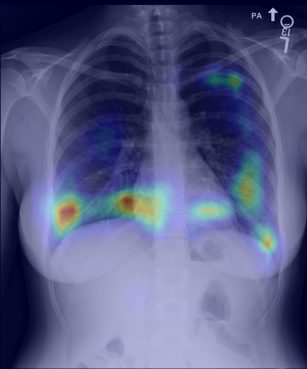}}
  \end{subfigure}}
\centering  \adjustbox{minipage=0.05\linewidth,valign=t}{
  \begin{picture}(0,120)
\put(-15,-0.4){\includegraphics[width=.238\linewidth,trim={0.45cm 0cm 0cm 0cm},clip]{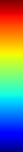}}
\put(-12,-2){-Min}
\put(-12,102){-Max}
\end{picture}
}}
  \caption{Examples of heatmaps over the respective CXR. a) ET map of a radiologist; b) average ET map of the remaining four radiologists; c) segmentation of the lung region used as a baseline d) average ET map from all CXRs, center bias (CB), after registration to match the location of the lungs; e) saliency map for a model without attention gates (woAG) generated by Grad-CAM with uniform weights; f) saliency map for a model with attention gates (wAG) generated by Grad-CAM with uniform weights; g) Attention map 1 (AM1) from wAG; h) Attention map 2 (AM2) from wAG. } \label{examples}
\end{figure}

We use the REFLACX dataset~\cite{reflacx,physionet,reflacxarxiv}, which focuses on \acp{CXR}, to build \ac{ET} maps from radiologists' gazes and compare them with saliency maps from abnormality classification models. Figure~\ref{examples} shows examples of \ac{ET} maps and generated saliency maps. Differences are expected between them. Whereas a radiologist looks at multiple locations to inspect for abnormalities, interpretability methods are expected to highlight the areas where changes would cause a large impact on the output, i.e., abnormalities.

There are reasons for human and model heatmaps to be similar. Grad-CAM, one of the most used explanation methods in the medical field~\cite{popular}, should provide low-resolution smooth saliency maps, similar to the \ac{ET} maps. When comparing to human heatmaps, Ebrahimpour et al.~\cite{attend} showed the superiority of the similar \ac{CAM}~\cite{cam} method. Since Grad-CAM provides one saliency map per class, we empirically test a few methods of combining them. Attention maps, self-explanatory masks that multiply spatial feature maps of a network, may be similar to \ac{ET} maps since they are intrinsically class-independent and because of their human attention inspiration~\cite{inspire}.

We calculate upper and lower bounds for similarity scores to better understand their expected range. Because of the proximity of the two bounds in traditional metrics, we use shuffled metrics to correct for center biases~\cite{centerbias}. There is a tendency for fixations, i.e., image locations gazed at by radiologists, to be in central regions of the images. Saliency maps concentrating on these regions achieve high scores independently of image content. Shuffled metrics try to fix this problem and are formulated so that differences and similarities between heatmaps have different weights on the final scores depending on how commonly gazed their locations are.  Given the structural similarity of \acp{CXR}, we calculate a specific center bias for this task, as shown in Figure~\ref{main:d}. Finally, we evaluate the generated saliency map, reaching scores comparable to the interobserver agreement in one of the metrics.

\subsection{Related work}

In the field of interpretability, a few works have used \ac{ET} maps to evaluate explanatory saliency maps~\cite{attend,sidu,iris}. Ebrahimpour et al.~\cite{attend} collected \ac{ET} maps from participants listing objects present in natural images. They compared the data against the saliency maps of the object-detection models for the class with the highest score. In the field of medical images, Trokielewicz et al.~\cite{iris} compared the Grad-CAM~\cite{gradcam} saliency maps against humans in the task of iris recognition. Muddamsetty et al.~\cite{sidu} did similar work for classification tasks of retinal images. These works analyzed only binary medical tasks and did not evaluate a strong baseline related to the biases present in \ac{ET} data. Karargyris et al.~\cite{ibm} qualitatively checked the Grad-CAM saliency maps against \ac{ET} maps in \acp{CXR}, with no quantitative analysis. To the best of our knowledge, our study is the first to perform this quantitative analysis on \acp{CXR}. 

The field of automatic generation of \ac{ET} maps uses \ac{ET} data as ground-truth and training data~\cite{centerbias}. We employ the same comparison metrics as this field, but we do not focus on generating a saliency map that best matches \ac{ET} maps.

\section{Methods}
\subsection{Grad-CAM}
Grad-CAM~\cite{gradcam} generates a saliency map according to
\begin{equation}
\label{eq:gradcam_general}
GC_c=ReLU\left(\sum_k{\alpha_c^kLSFM^k}\right),
\alpha_c^k=GAP\left(\frac{\partial logit(x)_c}{\partial LSFM^k}\right),
\end{equation}
where $GC_c$ is the saliency map for each class $c$, $ReLU$ is a rectified linear unit, $\alpha^k_c$ is a weight for channel $k$ of the last spatial layer of a network, $GAP$ is global average pooling, $logit(x)_c$ is the logit output for the model being evaluated for class $c$, and $LSFM^k$ are the activations for channel $k$ of the last spatial feature maps of a network. In other words, Grad-CAM calculates a combination of the \acp{LSFM} of a network and the gradient of the network outputs with respect to each element of the \acp{LSFM}. To combine the $GC_c$ from all classes, we use
\begin{equation}
\frac{1}{\sum_c{\psi_c}}\sum_c{\psi_c\times GC_c},
\end{equation}
where $\psi_c$ is a weight for the saliency map of each class. We consider three ways of choosing the weights $\psi_c$ to mix the $GC_c$ for each class $c$:
\begin{itemize}
\item \textbf{Thresholded}: uniformly mix the classes that are considered present in the image based on a threshold on the model's output, according to
\begin{equation}
\psi_c = \begin{cases}
    1,\text{if } logit(x)_c>0\\
    0,\text{if } logit(x)_c<0
  \end{cases}.
\end{equation}
If all $\psi_c$ are 0 for an image, we assign $\psi_c=1$ for the ``No Finding'' label.
\item \textbf{Weighted}: weight the classes using the output of the model, according to $\psi_c = \sigma(y_c)$, where $\sigma$ is the sigmoid function.
\item \textbf{Uniform}: uniformly mix all the classes: $\psi_c = 1$.
\end{itemize}

\begin{figure}
\center
\includegraphics[width=.7\linewidth]{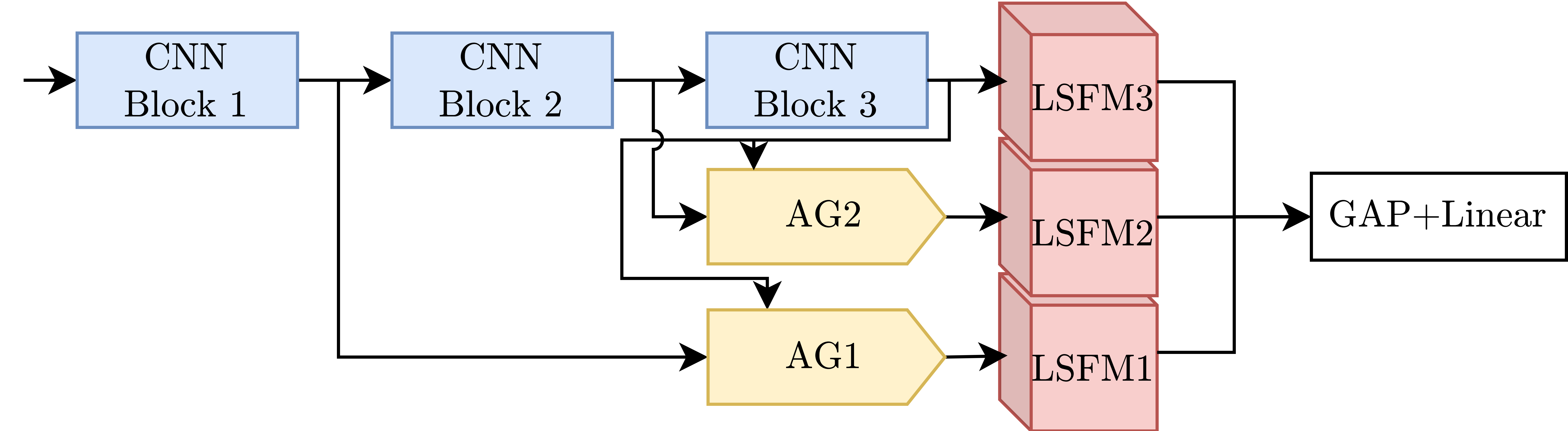}
\caption{Network with attention gates (wAG). AG stands for attention gate. The network without attention gates (woAG) has a similar architecture, with only LSFM3 input to the GAP operation. LSFM represents the activation maps used for calculating the Grad-CAM saliency map. The description of the CNN Blocks are given in Section~\ref{training}, and of the attention gates (AG) in Figure~\ref{fig:ag}.}
\label{fig:agsono}
\end{figure}

\begin{figure}
\center
\includegraphics[width=.65\linewidth]{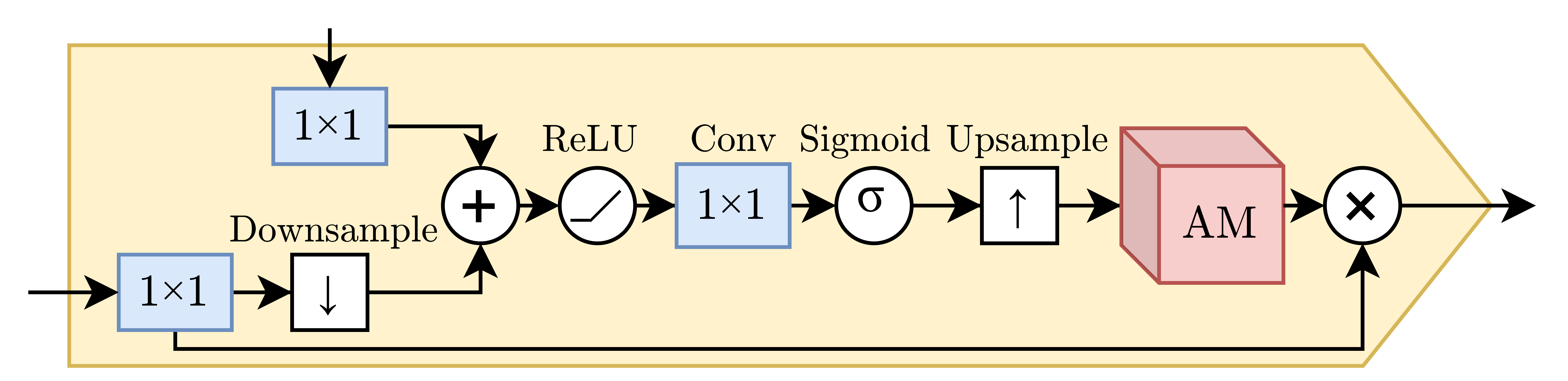}
\caption{Operations inside the attention gate (AG). AM stands for attention map.}
\label{fig:ag}
\end{figure}

\subsection{Attention gates}

Figure~\ref{fig:agsono} shows the architecture of the gated \ac{CNN}, including the location of the attention gates. Figure~\ref{fig:ag} shows the employed attention gates. Each attention gate provides a saliency map through its attention map, and the attention maps can also be combined into a single saliency map through the use of Grad-CAM. Each channel from the \acp{LSFM} on Figure~\ref{fig:agsono} is considered as one of the $k$ channels from Equation~\ref{eq:gradcam_general}. The output from CNN Block 3 in Figure~\ref{fig:agsono} is directed to the attention gates before being called LSFM3 so that the gradient calculation from Equation~\ref{eq:gradcam_general} is influenced by LSFM3 in only one of the three \ac{CNN} branches. 

\section{Experiments}

\label{exp}
Experiments employed \acp{CXR} from the MIMIC-CXR-JPG dataset~\cite{physionet,mimicxrjpgphysionet,mimiccxrjpg}, which contains \acp{CXR} patients admitted to the ER at the Beth Israel Deaconess Medical Center from 2011 to 2016, and \ac{ET} data from the REFLACX dataset~\cite{reflacx,physionet}, which contains data collected while radiologists dictated reports. The use of the datasets was exempt from approval since they are publicly available and de-identified. All experiments used a \ac{CNN} as depicted in Figure~\ref{fig:agsono}. We used the parts of the REFLACX dataset where the same \acp{CXR} contained \ac{ET} data for all five radiologists, totaling 91 \acp{CXR}. For each of the readings, one heatmap was generated from the fixations. We used PyTorch 1.8.1~\cite{pytorch} in our experiments.

\subsection{Classification models}
\label{training}
We trained two types of models, with (wAG) and without attention gates (woAG). They were trained and validated using the MIMIC-CXR-JPG dataset. We used the Adam optimizer~\cite{adam} with a learning rate of 0.0001 and a weight decay of $0.00001$. After three epochs without improvement on the validation average \ac{AUC}, we multiplied the learning rate by 0.5. We trained with the binary cross-entropy loss for 75 epochs and a batch size of 64. The 14 labels and the data split from the MIMIC-CXR-JPG dataset were used. All images from subjects displayed to radiologists were moved to the test set. The training was limited to images filtered as follows: images without classification labels were discarded; only frontal \acp{CXR} were kept, i.e., images with ``ViewPosition'' equals to ``AP'' (anterior-posterior) or ``PA'' (posterior-anterior); and studies with more than one frontal image were excluded. The \ac{CNN} blocks from Figure~\ref{fig:agsono} were built following the Sononet-16~\cite{sononet} architecture, but with the attention gate inclusion from Schlemper et al.~\cite{ag}.

For training, data were resized to have its shortest dimension equal to 224 pixels, rotated between -15 and +15 degrees, translated by up to 5\% of its dimensions, scaled with a scale factor between 0.95 and 1.05, center cropped, randomly horizontally flipped, and normalized by the average intensities and standard deviation of the ImageNet dataset. For validation, images were scaled such that their longest dimension was a multiple of 16, and their shortest dimension the closest to 224 while keeping the aspect ratio. The image was then padded to a square. Saliency maps were generated with this padded version of the image and then cropped to the original aspect ratio.

For each method, we trained five models to calculate the variability of results, which are reported with their 95\% confidence intervals and the number of samples used to calculate the confidence intervals. On the classification test set ($n= 5$ seeds), woAG had an \ac{AUC}, averaged over the 14 labels, of 0.774 [0.771,0.776], whereas wAG had an average \ac{AUC} of 0.769 [0.765,0.772].

\subsection{Metrics and baselines}

\Ac*{ET} maps were generated by drawing Gaussians centered in each fixation and combining them through a sum weighted by the fixation duration. Following Le Meur \& Baccino~\cite{onedegree}, the Gaussians had a standard deviation of 1 degree of visual angle in each axis to represent location uncertainties. To reduce the variability of the scores~\cite{average} and use the fact that \acp{CXR} had \ac{ET} maps for five radiologists, we calculated the metrics against the average \ac{ET} map of all combinations of four radiologists (1 vs. 4), which were considered the ground truth.

From the literature on automatic generation of human saliency maps~\cite{centerbias}, we selected two metrics to compare saliency maps: \ac{NCC}, i.e., Pearson's correlation coefficient, and the Borji formulation of \ac{AUC}~\cite{borjiauc}. The \ac{NCC} was directly calculated between \ac{ET} map and generated saliency map. The \ac{NCC} metric calculates the similarity of the saliency maps on the whole image by calculating the correlation between both signals, following

\begin{equation}
\text{NCC}\left(x_1,x_2\right)=\frac{1}{P-1}\sum{\frac{\left(x_1\left(p\right)-\mu_{x_1}\right)}{\sigma_{x_1}}\times\frac{\left(x_2\left(p\right)-\mu_{x_2}\right)}{ \sigma_{x_2}}},
\end{equation}
\noindent where $x_1$ and $x_2$ are the two images of same size, $x_i(p)$ is the value of image $x_i$ in pixel coordinate $p$, $P$ is the total number of pixels in one image, $\mu_{x_i}$ is the average value of $x_i$, and $\sigma_{x_i}$ is the unbiased standard deviation of $x_i$.

The AUC metric considers each pixel of a saliency map to represent the output of a binary classifier deciding if each pixel should contain human fixations or not~\cite{centerbias}. This binary classifier should have higher output values for locations associated with fixations from humans (positive locations --- \ac{ET} map) when compared to the rest of the locations (negative locations --- uniform heatmap). We calculate the \ac{AUC} by sampling locations from the heatmaps, which represent spatial probability distributions. The AUC metric calculates the probability of the binary classifier outputting a higher value for a positive location than for a negative location. By looking at the standard deviation of AUCs calculated with a varied number of samples, we found that sampling 1000 positive and 1000 negative samples was enough to have a sufficiently accurate AUC.

To have an upper bound for the metrics, we measured interobserver scores. To find a lower bound, we used an algorithm to segment the lungs and calculated the convex hull of the segmentations to include the mediastinum and the bilateral hemidiaphragms, as shown in Figure~\ref{main:c}. Scores for the baselines are presented in Table~\ref{tab:scoresbaselines}. Upper and lower bounds were practically the same for \ac{NCC}. This small range was probably caused by a strong bias toward having fixations around the lung area. Thus, using these metrics does not provide much information about our generated saliency maps other than their capacity to highlight the lungs. To correct this bias, we used shuffled metrics~\cite{centerbias}, which penalize models that output this center bias and reward models that can highlight other regions of the image that were fixated more than the average center bias~\cite{centerbias}. \Acp{CXR} are very similar in terms of what is structurally present in an image, making our center bias issue even worse than for natural images. We also corrected for the lung location variance, making the similarity with a center bias even stronger. Therefore, the impact of using shuffled metrics for \ac{CXR} images is expected to be higher than for natural images. 

We used bounding box annotations for lungs and heart to calculate the center bias in our dataset. We calculated the average bounding box, registered all bounding boxes to the average, applied the same transformation to the \ac{ET} map, and combined them to get an average of the fixations. The resulting center bias is shown in Figure~\ref{main:d}. For use in the metrics, the center bias is transformed to match the bounding box location of the respective \ac{CXR}.

For the \ac{NCC} calculation, we drew from a closely related metric~\cite{centerbias}, normalized scanpath saliency, from the shuffled formulation by Gide \& Karam~\cite{snss},
\begin{align*}
sNCC(GT,SM) =NCC(GT, SM)-NCC(CB, SM),
\end{align*}
where sNCC is the shuffled \ac{NCC}, GT is the ground truth saliency map, SM is the saliency map being evaluated, and CB is a heatmap representing the center bias in the dataset.

The only change from AUC to shuffled \ac{AUC} (sAUC) is in sampling negative locations from the center bias map instead of sampling them from the uniform heatmap. This change compensates for the center bias in the metric and assigns higher scores for models that focus on outputting saliency maps highlighting locations that are fixated more than average.

There are differences between the employed metrics. The AUC metric does not penalize false positives, making saliency maps that are blurrier than the ground truth not be penalized~\cite{centerbias}. The NCC metric is equally affected by false negatives and false positives. sAUC can be between 0 and 1, the higher, the better, and 0.5 represents a model with random outputs. sNCC can be between -2 and 2, the higher, the better, and 0 represents a model with random outputs. 

\begin{table}[ht]
\caption{Scores for the tested methods of generating saliency maps (SM). We highlight in bold the highest-scoring saliency map for each metric, excluding the interobserver upper bound. Confidence intervals were calculated using $n=91$ CXRs.}
\label{tab:scoresbaselines}
\begin{center}
\begin{small}
\begin{sc}
\center
\begin{tabular}{|l|l|c|c|}
\hline
 SM	&	Method	& sNCC	& sAUC	\\ \hline
Interobs.	&	Baseline	&	0.632 [0.606,0.658]	&	0.790 [0.782,0.799]		\\
Segment.	& Baseline	&	\bm{$0.637~[0.615,0.659]$}	&	\bm{$0.735~[0.726,0.745]$}	\\  \hline
G-CAM	&	Thresh.	&	0.252 [0.200,0.304]	&	0.596 [0.574,0.619]		\\
(woAG)	&	Weighted	&	0.408 [0.368,0.447]	&	0.683 [0.667,0.699]		\\
	&	Uniform	&	0.437 [0.403,0.471]	&	0.696 [0.682,0.711]	\\
G-CAM	&	Thresh.	&	0.194 [0.158,0.229]	&	0.583 [0.568,0.598]	\\
(wAG)	&	Weighted	&	0.299 [0.263,0.335]	&	0.672 [0.659,0.685]	\\
	&	Uniform	&	0.343 [0.311,0.375]	&	0.678 [0.665,0.692]	\\
AM1	&	-	&	0.254 [0.225,0.284]	&	0.672 [0.657,0.688]	\\
AM2	&	-	&	0.359 [0.326,0.392]	&	0.684 [0.671,0.697]	\\
\hline
\end{tabular}
\end{sc}
\end{small}
\end{center}

\end{table}

\begin{table}[ht]
\caption{Shuffled scores for the tested methods of generating saliency maps (SM). We highlight in bold the highest-scoring saliency map for each metric, excluding the interobserver upper bound. Confidence intervals were calculated using $n=91$ CXRs.}
\label{tab:scores}

\begin{center}
\begin{small}
\begin{sc}
\center

\begin{tabular}{|l|l|c|c|}
\hline
 SM	&	Method	&	 sNCC	& sAUC	\\ \hline
Interobs.	&	Baseline	&	\,\,0.028 [\,\,0.002,\,\,0.055]	&	0.558 [0.547,0.568]	\\
Segment.	& Baseline	&	-0.187 [-0.204,-0.169]	&	0.505 [0.500,0.510]	\\ \hline
G-CAM	&	Thresh.	&	-0.035 [-0.058,-0.012]	&	0.510 [0.501,0.519]	\\
(woAG)	&	Weighted	&	-0.060 [-0.082,-0.038]	&	0.521 [0.511,0.530]	\\
	&	Uniform	&	-0.067 [-0.089,-0.046]	&	0.522 [0.513,0.532]	\\
G-CAM	&	Thresh.	&	-0.002 [-0.024,\,\,0.019]	&	0.512 [0.504,0.519]	\\
(wAG)	&	Weighted	&	\,\,0.027 [\,\,0.002,\,\,0.053]	&	0.528 [0.519,0.537]	\\
	&	Uniform	& \bm{$\,\,0.029~[\,\,0.002,\,\,0.055]$}	& \bm{$0.529~[0.521,0.538]$}	\\
AM1	&	-	&		-0.032 [-0.055,-0.009]	&	0.514 [0.505,0.523]	\\
AM2	&	-	&		-0.007 [-0.034,\,\,0.020]	&	0.522 [0.511,0.533]	\\
\hline
\end{tabular}

\end{sc}
\end{small}
\end{center}

\end{table}

The results of the shuffled metrics for the baselines are presented in Table~\ref{tab:scores}. The slightly positive value of the sNCC metric shows that the \ac{ET} map from each radiologist is slightly more similar to the average of the other radiologists than to the center bias. The more extensive range between baseline bounds shows that considering the center bias is essential for calculating a meaningful score.

\subsection{Results and discussion}
\label{sec:results}
Table~\ref{tab:scores} reports the metrics for all the tested methods. Not considering the baselines, the woAG model had the highest scores for the non-shuffled metrics and the wAG model for the shuffled metrics. Grad-CAM with uniform $\psi_c$ had the highest score in both cases. Uniform $\psi_c$ might have achieved the best results because radiologists have to look for all abnormalities, including those not found in a particular image. Considering the sNCC metric, one of the models reached scores almost identical to the interobserver evaluation. For the AUC metrics, the interobserver evaluations had the highest scores with a good margin, highlighting that each metric measures different qualities of the heatmaps. Although the attention maps were not the highest scoring saliency maps, the Grad-CAM method had the highest shuffled scores when applied to the wAG model, showing an advantage of attention-gated models when compared to human attention.

We also analyzed scores splitting normal and abnormal cases. Abnormal cases had a majority of radiologists selecting at least one abnormality for the image.
As shown in Table~\ref{tab:abnormal}, interobserver scores and the scores from a chosen interpretability method were higher for abnormal CXRs. This difference might have been caused by normal cases not having an evident area of interest and abnormalities being areas of longer fixations by radiologists and stronger saliency for Grad-CAM. The scores for the segmentation baseline showed almost no change.

\begin{table}
\caption{Scores of baselines and of the Grad-CAM (wAG) with uniform $\psi_c$ method when splitting the dataset into normal (N) and abnormal (Abn) CXRs. Confidence intervals were calculated using $n=17$ normal CXRs and $n=74$ abnormal CXRs.}
\label{tab:abnormal}
\begin{center}
\begin{small}
\begin{sc}
\center
\begin{tabular}{|l|l|c|c|c|c|c|c|} \hline
   Metric &  LBL &  Interobserver (IO) &  Segmentation &  Grad-CAM (wAG) \\ \hline
sNCC	&	N	&	-0.056 [-0.110,-0.003]	&	-0.173 [-0.206,-0.141]	&	-0.047 [-0.096,0.002]	\\
sNCC	&	Abn	&	\,\,0.048 [\,\,0.020,\,\,0.076]	&	-0.191 [-0.212,-0.171]	&	\,\,0.045 [\,\,0.016,0.074]	\\
sAUC	&	N	&	\,\,0.532 [\,\,0.511,\,\,0.553]	&	\,\,0.504 [\,\,0.497,\,\,0.512]	&	\,\,0.497 [\,\,0.482,0.512]	\\
sAUC	&	Abn	&	\,\,0.566 [\,\,0.554,\,\,0.577]	&	\,\,0.507 [\,\,0.501,\,\,0.513]	&	\,\,0.538 [\,\,0.529,0.548]	\\
\hline
\end{tabular}
\end{sc}
\end{small}
\end{center}
\end{table}


\section{Conclusion}
Using a dataset of \ac{ET} data from five radiologists, we showed that, when controlling for center bias, interpretability maps can be as similar to the \ac{ET} maps from radiologists as \ac{ET} maps from other radiologists. In other words, although the tested saliency maps are not good at highlighting areas fixated regularly in the average CXR, they excel at highlighting the specific areas in each CXR that radiologists fixate more than average. In our evaluation, the Grad-CAM method with uniformly weighted saliency maps of each class produced maps more similar to the \ac{ET} maps. The attention-gated model produced saliency maps with the highest scores, showing an advantage of human-inspired spatial attention gates for generating heatmaps similar to human attention. Moreover, higher similarity scores were associated with the presence of abnormalities, which is similar to our expectations, given that spatial heatmaps are usually formulated to highlight the presence of a class.


%
%
%
\bibliographystyle{splncs04}
\bibliography{references_short}

\end{document}